# BELL: Benchmarking the Explainability of Large Language Models

Syed Ahmed

syed_ahmed12@infosys.com

Karthick Selvaraj

karthick.s50@infosys.com

Bharathi Vokkaliga Ganesh

bharathi_ganesh01@infosys.com

ReddySiva Naga Parvathi Devi

reddy.devi@infosys.com

Jagadish Babu P

jagadish.p02@infosys.com

Sravya Kappala

kappala.sravya@infosys.com

Responsible AI Office
Infosys Limited, Bangalore, India

**Abstract:** Large Language Models (LLMs) have demonstrated remarkable capabilities in natural language processing (NLP), yet their decision-making processes often lack transparency. This opaqueness raises significant concerns regarding trust, bias, and model performance. To address these issues, understanding and evaluating the interpretability of LLMs is crucial. This paper introduces a standardized benchmarking technique, **BELL (Benchmarking the Explainability of Large Language Models)**, designed to evaluate the explainability of large language models. To assess the reasoning capabilities of LLMs, a variety of thought-eliciting techniques are utilized, including Chain-of-Thought (CoT), Thread-of-Thought (ThoT), ReRead CoT, ReRead ThoT, Chain-of-Verifications (CoVe), Hallucination, Graph-of-Thought (GoT), and Logic-of-Thought (LoT). These evaluations are conducted using open-orca dataset, with the quality of explanations measured through a comprehensive set of metrics. These metrics encompass factors such as coherence, uncertainty, hallucination and cosine similarity relative to the model's internal reasoning. By systematically evaluating the performance of various LLMs against this benchmark, the goal is to identify models that offer more transparent and reliable explanations. This benchmarking technique serves as a valuable tool for researchers and practitioners in selecting appropriate LLMs for tasks that demand interpretability and accountability. To foster reproducibility and facilitate wider adoption, we have implemented each of the proposed and utilized explainability techniques as open-source software, with the code and necessary resources available at https://github.com/Infosys/Infosys-Responsible-AI-Toolkit.

## 1. Introduction

Large language models have revolutionized natural language processing and generative Artificial Intelligence (AI), as shown by numerous foundational studies [1]. These models' exceptional capabilities have attracted significant attention, enabling a wide range of applications. LLMs are utilized for tasks such as translation [2], content generation, content summarization, article writing [3], as well as enhancing search functions (Bing Chat [4]) etc., The impact of LLMs extends to fields like software development, with models like Code Llama [5] aiding engineers. Their applications also span finance sector [6], scientific research [7] [8], including areas such as arts [9], education [10], oceanography [11], law [12], political science [13], medicine [14] [15], showcasing their broad and diverse influence.

However, the exponential rise in use of LLMs also brings challenges related to their explainability and interpretability. First, the complexity of LLMs, driven by their diverse outputs and generative capabilities, makes it difficult to understand how they arrive at specific responses. This lack of transparency can lead to unpredictable [16] and potentially misleading outputs, complicating efforts to ensure trust in their performance. Second, biases [17] in training datasets can affect the fairness and interpretability of LLM outputs. For example, a model might favor certain perspectives due to data biases, making it difficult to explain or justify certain outputs. Third, the inclusion of sensitive data in training sets can compromise the interpretability of LLMs in ensuring privacy [18]. Additionally, high user expectations for accurate and human-aligned responses highlight the need for LLMs to be interpretable in ways that align with human values, as misalignment can lead to ethical concerns and reduce trust in their applications.

To tackle these challenges, it is essential to address the core issue of benchmarking the explainability and interpretability of LLMs. How can the transparency of their decision-making processes be systematically evaluated? Establishing robust benchmarking like **BELL** enables researchers to critically evaluate and compare LLMs, fostering the development of models that prioritize transparency and accountability. This approach not only addresses current concerns but also sets the foundation for future advancements in trustworthy AI systems. Key contributions of this paper include:

- **Standardized Benchmarking:** The paper introduces standardized benchmarking, **BELL**, employing diverse thought-eliciting techniques such as CoT, ThoT, GoT, ReRead, LoT, CoVe, and Hallucination to systematically evaluate the reasoning capabilities of LLMs. These methods provide a structured approach to assess the interpretability and transparency of model decision-making.
- **Metrics:** This paper describes a set of evaluation metrics, including coherence, uncertainty, and cosine similarity, to quantitatively measure the quality and transparency of LLM explanations.

A key contribution of this work is the public release of our implementation of explainability techniques and respective evaluation metrics enabling the community to readily adopt and extend our findings. The codebase is accessible at [Infosys-Responsible-AI-Toolkit/responsible-ai-llm-explain at master · Infosys/Infosys-Responsible-AI-Toolkit](Infosys-Responsible-AI-Toolkit/responsible-ai-llm-explain_at_master_·_Infosys/Infosys-Responsible-AI-Toolkit)

The structure of the paper is as follows: Section 2 presents the background and related work, while Section 3 introduces proposed benchmarking designed to evaluate the explainability of LLMs. Section 4 details the datasets and evaluation metrics, and Section 5 outlines the findings of the study. Section 6, which is the last section, concludes with insights and directions for future research.

## 2. Background

### 2.1 Interpretability

The rise of highly effective closed-source LLMs, such as ChatGPT, has sparked significant interest in these models. At the same time, open-source alternatives like Granite [19], Llama [20] have gained notable popularity. The growing adoption of LLMs has also captured the attention of researchers focused on model interpretability and explainability. Understanding how LLM capabilities evolve during training is critical to analyzing their formation and functionality in downstream tasks. Additionally, investigating inference processes and the influence of context is vital for uncovering insights into model behavior. However, this analysis is challenging due to the vast number of parameters and complex non-linear structures inherent to LLMs.

Most research on explainability and interpretability focuses on LLMs at a macroscopic level, analyzing their behaviors qualitatively. Many studies in this area utilize prompts to explore these aspects [21] [22].

Nonetheless, a smaller subset of works explores the inner workings of LLMs using probing methods such as prompt tuning [23] or activation analysis. In contrast, mechanical interpretability aims to develop theories based on the fundamental behaviors of transformers. These approaches focus on examining the core components that drive LLM creation and analyze their internal representations by injecting or training specific modules. However, applying probing or mechanistic interpretability techniques to closed foundation models presents significant challenges, as their internal architectures and parameters are often not accessible. In such cases, exploring thought-eliciting techniques offers an alternative, providing a means to validate the reasoning capabilities of models, regardless of whether they are open-source or closed-source.

## 2.2 Reasoning

Reasoning in large language models (LLMs) encompasses their ability to logically analyze information, address problems, and formulate conclusions based on the provided context. This capability enables LLMs to perform tasks that demand critical thinking, such as drawing inferences, identifying relationships, and making decisions. Techniques like Chain-of-Thought (CoT) prompting, Thread-of-Thought (ThoT) methodologies, Graph-of-Thoughts (GoT) and other frameworks [24-30] are widely employed to enhance reasoning capacities. These approaches promote the decomposition of complex problems into smaller, intermediate steps, thereby making the reasoning process more structured and efficient.

The integration of reasoning significantly enhances the explainability of LLMs by clarifying their decision-making processes. By explicitly detailing the steps leading to a conclusion, reasoning ensures that the model's output becomes more transparent and comprehensible. This clarity enables users to understand the logic behind the results, facilitating a more thorough evaluation of the model's reliability and accuracy. Moreover, the structured reasoning process bolsters confidence in the system by demonstrating how outcomes are derived.

Furthermore, reasoning fosters seamless collaboration between humans and AI. By aligning computational problem-solving strategies with human thought processes, LLMs become more intuitive and easier to interact with. This alignment enhances user trust and reliance, thereby improving the practical applicability and effectiveness of AI systems in diverse real-world scenarios.

The advancements in reasoning frameworks such as CoT, ThoT, and GoT demonstrate significant strides in enhancing LLMs logical reasoning capabilities. However, despite these innovations, a critical gap remains in the benchmarking of LLMs for explainability. Current benchmarks often emphasize performance metrics while overlooking the need for consistent, interpretable reasoning processes. This limitation becomes particularly concerning in tasks where trust and transparency are essential, as unfaithful reasoning can undermine the reliability of LLM outputs. Addressing this drawback, our work focuses on benchmarking LLMs for explainability, aiming to evaluate and improve their ability to provide coherent and transparent reasoning aligned with their predictions.

## 3. Benchmarking Explainability of Large Language Models (BELL)

Thought Eliciting Techniques in Large Language Models are strategies used to organize and direct the model's reasoning process, helping it produce responses that are logical, coherent, and focused. These methods emulate human problem-solving and critical thinking by simplifying complex questions into smaller steps, examining different possibilities, and improving clarity.

In this research, we investigate how employing thought-eliciting strategies enhances the ability of large language models to handle complex reasoning tasks effectively. We demonstrate that such reasoning capabilities naturally emerge in large models using straightforward prompt engineering techniques, including Chain-of-Thought, Thread-of-Thought, Graph-of-Thought, Logic-of-Thought, ReRead, and Chain-of-Verifications.

Fig-1 illustrates a comprehensive architecture for leveraging LLMs to enhance reasoning and evaluation processes. The process begins with input, comprising dataset and an LLM which serves as the foundation for subsequent stages. The "Technique" & "Thought-Elicitation" phase encompasses various reasoning techniques (discussed from section 3.1 to 3.2), including CoT, ThoT, ReRead CoT, ReRead Thot, CoVe, Hallucination. These components are guided by a prompt template to generate structured explanations. The outputs of this stage undergo evaluation based on metrics (discussed in section 4.2) such as cosine similarity, uncertainty quantification, coherence and hallucination. These evaluations collectively determine the final score, which represents the system's output, ensuring a robust and methodical approach to reasoning and decision-making tasks. This architecture highlights the integration of diverse reasoning methods with quantitative evaluation to assess the reasoning capabilities of LLMs in complex problem-solving.

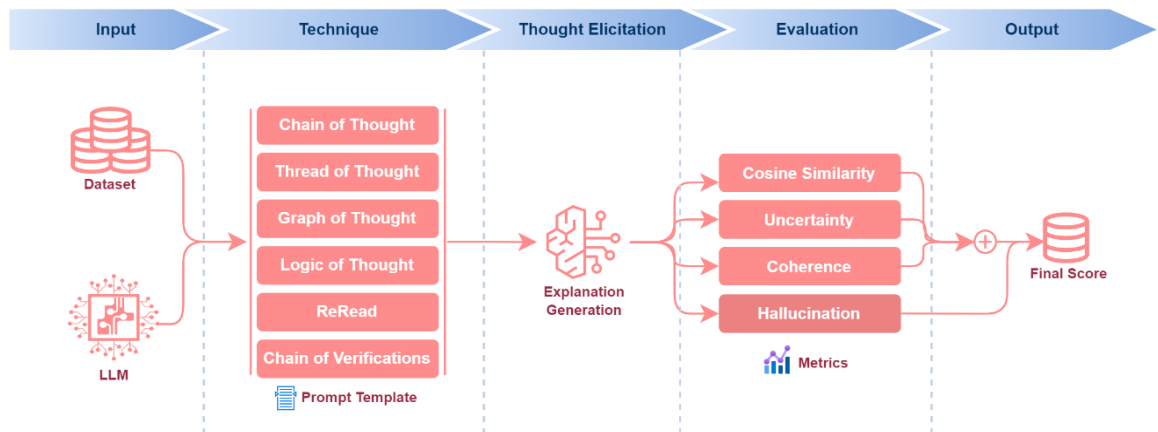

Fig 1: Proposed architecture

### 3.1 Chain of Thought (CoT):

Chain of Thought (CoT) reasoning, introduced by Wei et al. [31], is a process where each step logically follows the previous one, forming a structured path to a solution. This method helps LLMs tackle complex problems by breaking them down into smaller steps, similar to how humans connect ideas in a step-by-step manner. For example, when solving a math problem, the model first defines the variables, then applies the relevant equations, and finally computes the result step by step. The CoT approach has shown to significantly improve model accuracy, particularly when the reasoning process involves multiple stages, as it allows the model to verify and refine its thought process at each step. This approach has proven valuable in increasing the capabilities of large models like GPT-3 in reasoning-intensive tasks, highlighting how such models can be prompted to generate detailed, step-by-step explanations.

### 3.2 Thread of Thoughts (ThoT):

Thread of Thought (ThoT) by Zhou et al. [32], is a novel reasoning framework designed to improve the logical coherence and consistency of large language models in complex tasks involving multiple interconnected pieces of information. For example, when analyzing a lengthy, disorganized document, the model is prompted to break down the text into manageable parts, examine each segment step by

step - summarizing and analyzing as it proceeds - and then synthesize the insights to form a coherent understanding. This technique is particularly beneficial for tasks that require long-term reasoning or involve ambiguous or chaotic contexts, as it enables the model to adapt its thinking process as new information is introduced. ThoT reduces the risk of logical errors or inconsistencies that may arise when reasoning steps are treated in isolation, making the model's responses more reliable and interpretable.

### 3.3 ReRead

ReRead by Xu et al. [35], improves reasoning in large language models by incorporating a "re-reading" process, where the model revisits its prompt. This technique aims to enhance the model's ability to correct errors and adjust its reasoning based on a deeper understanding of the context. By prompting the model to re-examine its prompt, ReRead allows it to resolve ambiguities and inconsistencies. For instance, when faced with a complex problem, the model first reads the question to grasp the general context and then re-reads it to capture finer details and nuances. This dual-pass approach allows the model to refine its comprehension, leading to more accurate and coherent responses. Xu [35] demonstrate that this iterative approach significantly improves the model's performance on complex tasks, particularly those requiring multi-step reasoning or clarification of ambiguous queries. Further analyses show that with its adaptability, it can be easily integrated with any other thought eliciting technique to get better result. We have combined this with CoT & Thot to get different improved explanation and benchmarked the same.

### 3.4 Chain of Verification (CoVe):

Chain-of-Verification (CoVe) by Dhuliawala et al. [36], addresses the issue of hallucination in large language models, where models generate plausible but incorrect information. CoVe introduces a multi-step process where the model first drafts an initial response and then generates verification questions to fact-check its output. The model independently answers these questions to avoid bias from the initial response, followed by the generation of a final verified answer. This method helps correct errors by allowing the model to deliberate on its reasoning and ensure its accuracy.

### 3.5 Graph of Thought (GoT):

Graph-of-Thought (GoT) by Besta et al. [33], enhances large language models reasoning by organizing the thought process into a graph structure. In this framework, each reasoning step is represented as a node, and the relationships between steps are captured by edges, allowing for clear tracking of dependencies and logical flow. For instance, when planning a trip, GoT can explore different date options while also considering how each date impacts potential accommodation and activity choices, creating a network of interconnected decisions. This allows for more flexible and context-aware problem-solving compared to a simple chain of reasoning. GoT improves the model's ability to reason coherently and generate accurate responses. It also enhances interpretability, as the graph structure provides a transparent view of the model's reasoning path.

### 3.6 Logic of Thought (LoT):

Logic of Thought (LoT) by Tongxuan et al. [34], integrates logical reasoning directly into the contextual understanding of large language models. This approach focuses on explicitly infusing logical principles, such as deductive and inductive reasoning, into the model's decision-making process, allowing it to perform more structured and accurate reasoning. By embedding logic within the context of the problem, LoT enhances the model's ability to draw reliable conclusions, particularly in tasks requiring formal reasoning or complex problem-solving. The method also enhances interpretability by making the reasoning steps transparent and logically grounded.

# 4 Experimental Setup

The experiments were conducted on a 64-bit Windows 10 operating system utilizing an Nvidia NC16as_T4_v3 GPU configuration with 16 GB of dedicated GPU memory. The experimental environment was configured with Python, along with the requisite libraries and dependencies, to support the implementation and execution of transformer-based models.

## 4.6 Dataset

OpenOrca dataset [37] is an augmented version of the FLAN Collection, consisting of approximately 1 million completions from GPT-4 and 3.2 million completions from GPT-3.5. Each entry includes a question from the FLAN collection, submitted to either GPT-4 or GPT-3.5, with the corresponding response recorded. This dataset is suitable for tasks such as language modeling, text generation, and text augmentation, making it a valuable resource for developing and evaluating generative AI models for reasoning.

The dataset encompasses a diverse range of categories such as, mathematical problem-solving, sentiment analysis, etc. For the purpose of this research, the evaluation was specifically conducted on the mathematical problem-solving category.

## 4.7 Quality Evaluation Metrics

- **Cosine Similarity:** Cosine Similarity is a method used to evaluate how closely a generated response aligns with a base or reference response. Both generated and reference responses are transformed into vector representations in the model's embedding space, and cosine similarity is calculated to measures the degree of semantic similarity between them. This technique helps assess whether the model's output captures the intended meaning or context of the reference, making it valuable for evaluating performance in tasks like text generation, summarization, and question answering using eq. (1).

$$Cosine\ Similarity = \frac{A.B}{||A||\ ||B||} \quad (1)$$

Here, $A$ and $B$ are the embeddings of the generated and reference responses, $A.B$ is their dot product, $||A||$ and $||B||$ are their magnitudes calculated as shown below.

$$||A|| = \sqrt{\sum_{i=1}^{n} A_i^2},\ ||B|| = \sqrt{\sum_{i=1}^{n} B_i^2}$$

where, $A_i$ and $B_i$ are the components of the vectors $A$ and $B$, respectively, and $n$ is the dimensionality of the vectors.

- **Uncertainty:** Uncertainty indicates how confident the model is in its predictions or generated responses. It highlights situations where the model might lack clarity or face ambiguity in understanding an input. For instance, uncertainty is higher with incomplete or vague queries. By assessing uncertainty, it becomes easier to identify responses that may need validation or refinement. This metric is crucial for improving model reliability and performance. It also helps in tasks like filtering low-confidence outputs or integrating human-in-the-loop systems for better decision-making.
- **Coherence:** Coherence refers to how logically consistent and contextually aligned a generated response is with the input or surrounding text. It measures the flow of ideas and the internal consistency of the response. A coherent response stays on topic and maintains a clear, structured narrative throughout. Coherence ensures that the model's output makes sense and remains relevant to the input, improving the quality of text generation, summarization, and dialogue. High coherence is essential for generating human-like and contextually appropriate responses.

- **Hallucination:** Hallucination in Large Language Models refers to the phenomenon where the model generates outputs that are factually incorrect, nonsensical, or fabricated, even though they may sound plausible or grammatically correct. With the help of G-Eval metrics and cosine similarity of orca dataset, hallucination score can be detected from the LLM response.
  Hallucination Score = 1 - (0.8 * Average of Evaluation Metrics) - (0.2 * Average Similarity Score)

# 5 Findings

The performance evaluation of language model's explainability is conducted through a detailed assessment process. This involves calculating key metrics such as coherence and uncertainty for each generated response to measure the logical consistency and confidence in the outputs. Additionally, the cosine similarity between each response and a predefined baseline response is computed to evaluate the alignment with expected outcomes. For a comprehensive evaluation, these individual metrics are averaged to derive a combined score for each response. Finally, an overall explainability final score for the model is obtained by taking the average across all responses in the dataset. This approach provides a systematic and quantitative measure to assess the explainability of language models effectively. The evaluations results are illustrated from Graph 1 - Graph 7.

- (D) = The dataset of responses.
- (R_i) = The (i)-th response in the dataset (D).
- (E_i) = The generated explanation for the (i)-th response (R_i)
- (B_i) = The predefined baseline response for the (i)-th response (R_i).
- ({Coherence}(E_i)) = The coherence score of the explanation related to response (R_i).
- ({Uncertainty}(E_i)) = The uncertainty score of the explanation related to response (R_i).
- ({CosSim}(E_i, B_i)) = The cosine similarity score between the explanation (E_i) and the baseline response (B_i).
- (n) = The total number of responses in the dataset (D).
- ({OverallScore}) = The final overall explainability score for the given eliciting technique.
- {OverallScore} = frac{1}{n} \sum_{i=1}^{n} \left( \frac{{Coherence}(E_i) + {Uncertainty}(E_i) + {CosSim}(E_i, B_i)}{3} \right)
- ({Hallucination}(E_i)) = The hallucination score of the explanation related to response (R_i).
- ({Model_Score}) = Avg ({OverallScore}) - ({Hallucination}(E_i))

| Model | CoT | ThoT | ReRead CoT | Reread ThoT | CoVe | Hallucination | Model Score |
|---|---|---|---|---|---|---|---|
| GPT-4 | 85.28 | 92.39 | 91.91 | 91.37 | 85.14 | 19.42 | 87.78 |
| Gemma-2 9B | 72.91 | 91.73 | 90.41 | 91.8 | 83.08 | 25.07 | 84.15 |
| Mistral 7B | 79.93 | 90.34 | 79.94 | 89.71 | 88.33 | 26.4 | 83.64 |
| Llama-3.2 3B | 76.95 | 88.7 | 88.76 | 86.8 | 82.21 | 31.96 | 81.91 |
| Llava-1.6 7B | 58.81 | 86.87 | 85.93 | 87.95 | 82.13 | 25.1 | 79.43 |
| Nemotron-mini-4B-instruct | 79.79 | 75.44 | 74.94 | 78.37 | 78.46 | 26.4 | 76.95 |
| Llama-3.2 1B | 75.7 | 83.19 | 84.03 | 83.19 | 67.51 | 34.33 | 76.55 |

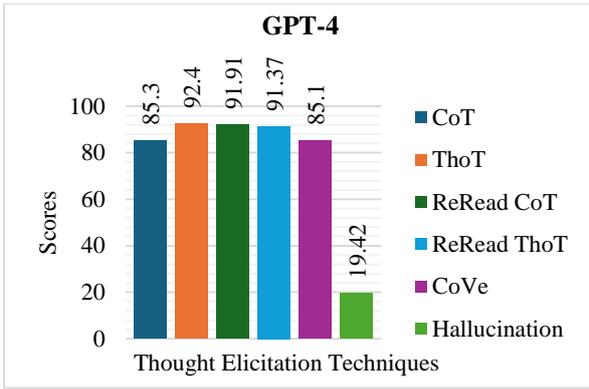
Graph 1: Evaluation of GPT-4

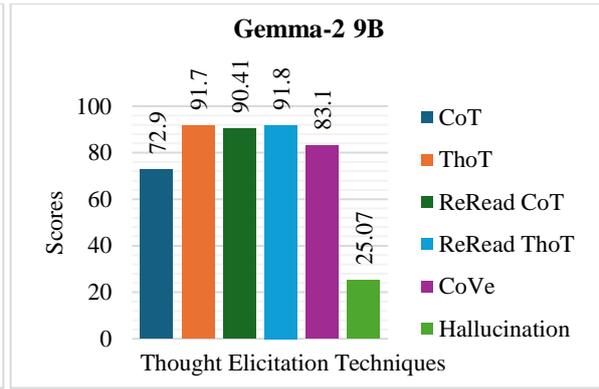
Graph 2: Evaluation of Gemma-2 9B

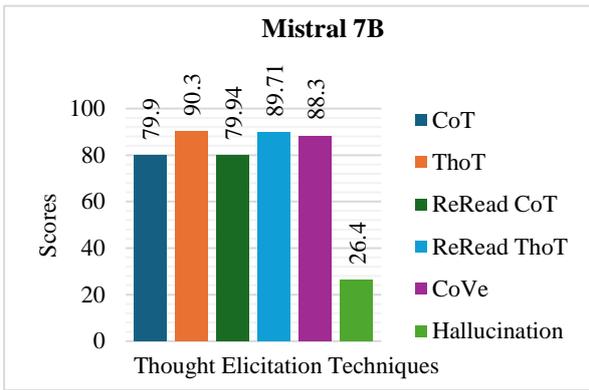
Graph 3: Evaluation of Mistral 7B

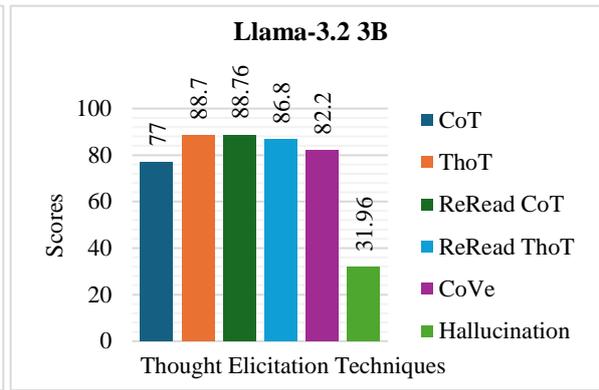
Graph 4: Evaluation of Llama-3.2 3B

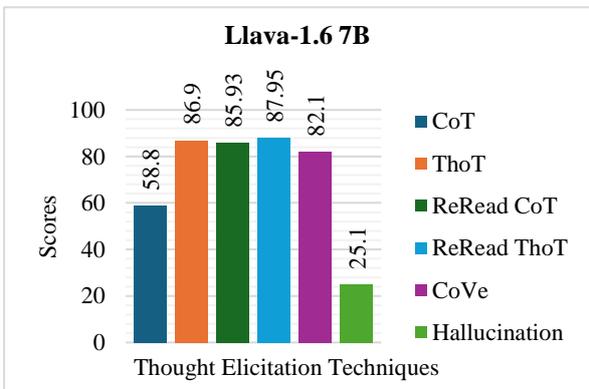
Graph 5: Evaluation of Llava-1.6 7B

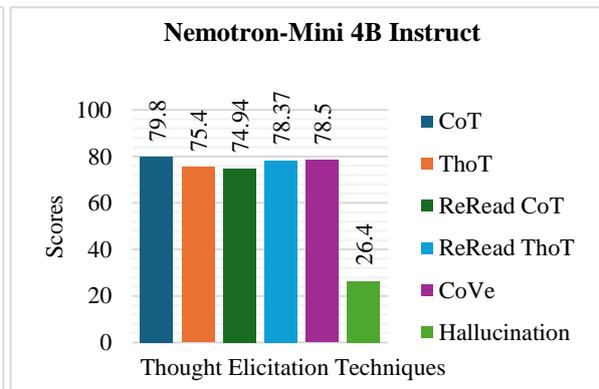
Graph 6: Evaluation of Nemotron-Mini-4B-Instruct

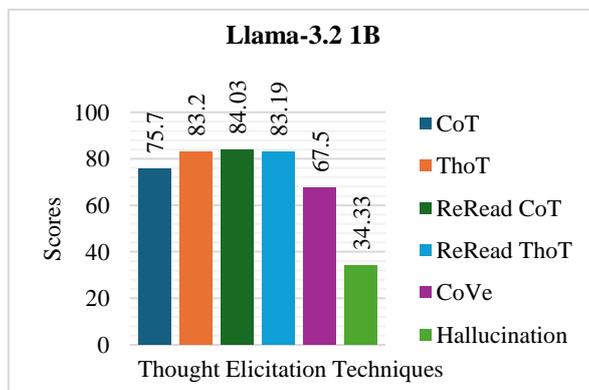
Graph 7: Evaluation of Llama-3.2 1B

In this study, we have assessed reasoning techniques, including Chain of Thought, Thread of Thought, ReRead CoT, ReRead ThoT, Chain of Verification, and Hallucination, across multiple models using cosine similarity, uncertainty and coherence. Among the models evaluated, GPT-4 consistently outperformed others, demonstrating robust sequential reasoning with a CoT score of 85.28 and leading performance in ThoT (92.39) and ReRead techniques. Gemma-2 9B and Mistral 7B followed closely in several categories, though they displayed minor gaps in reasoning and hallucination control compared to GPT-4.

Smaller models such as Llama-3.2 1B exhibited notable limitations, particularly in CoVe and hallucination scores, where their reliability was significantly lower than the larger counterparts that are evaluated. Despite moderate performance in ThoT and ReRead techniques, models like Llava-1.6 7B and Nemotron-mini-4B-instruct struggled with coherence and verification tasks, reflecting their constrained reasoning capabilities. Hallucination scores further highlighted discrepancies, with GPT-4 maintaining the lowest score of 19.42, while smaller models faced challenges exceeding acceptable thresholds.

Overall, these findings emphasize the superiority of larger models in reasoning tasks and the need for significant improvements in smaller architectures.

# 6   Conclusion and Future Work

The findings presented herein underscore the multifaceted nature of explainability. There is no single "best" XAI technique; rather, the optimal choice depends heavily on the specific application domain, the characteristics of the AI model being explained, and the intended audience of the explanations. This study evaluated the performance of several reasoning techniques across various models, revealing that larger models, such as GPT-4, consistently outperformed smaller ones. This is demonstrated using reasoning techniques such as, Chain of Thought (CoT), Thread of Thought (ThoT), ReRead techniques, and Chain of Verification (CoVe). GPT-4 demonstrated superior sequential reasoning and minimal hallucination, establishing its dominance in reasoning tasks. Conversely, smaller models, including Llama-3.2 1B and Llava-1.6 7B, displayed significant limitations in coherence and verification, underscoring the need for targeted improvements in their reasoning capabilities. The results emphasize that while smaller model's performance is moderate, larger models remain more reliable for complex reasoning tasks.

Future research explores a broader range of models, particularly with latest and emerging architectures that offer enhancements in reasoning capabilities. Additionally, we plan to assess the model's reasoning capabilities with latest eliciting techniques such as Graph of Thought, Logic of Thought using diverse datasets from fields like healthcare, law, and scientific research to understand how well models generalize across different domains.

As researchers, our aim in this white paper has been to lay a foundational framework for the rigorous evaluation and comparison of Explainable AI (XAI) techniques. We have explored a range of methodologies, metrics, and considerations crucial for understanding the strengths, weaknesses, and applicability of various XAI approaches. Our benchmarking efforts, while providing initial insights, serve as a steppingstone in the ongoing journey towards building truly transparent and trustworthy artificial intelligence systems. We strongly encourage readers – fellow researchers, practitioners, and policymakers – to leverage the insights and methodologies presented in this paper in their own work. Specifically, we urge you to leverage our code that is available here at [Infosys-Responsible-AI-Toolkit/responsible-ai-llm-explain at master · Infosys/Infosys-Responsible-AI-Toolkit](#) and extend the work presented here by developing new explainability techniques & evaluation metrics, curating diverse and challenging benchmark datasets, and evaluating a wider range of emerging XAI techniques.

## Abbreviations:

- BELL - Benchmarking the Explainability of Large language model
- LLM - Large Language Model
- NLP – Natural Language Processing
- CoT – Chain of Thought
- ThoT – Thread of Thought
- GoT – Graph of Thought
- LoT – Logic of Thought
- CoVe – Chain of Verification
- GPT - Generative Pre-trained Transformer.
- GPU - Graphics Processing Unit
- FLAN - Fine-tuned LAnguage Net